# Applying Incremental Deep Neural Networks-based Posture Recognition Model for Injury Risk Assessment in Construction

Junqi Zhao[a], Esther Obonyo[a,b]

a. Department of Architectural Engineering, The Pennsylvania State University, State College, PA, 16802, U.S.
b. Engineering Design, Technology, and Professional Programs, The Pennsylvania State University, State College, PA, 16802, U.S.

**Abstract.** Monitoring awkward postures is a proactive prevention for Musculoskeletal Disorders (MSDs) in construction. Machine Learning (ML) models have shown promising results for posture recognition from Wearable Sensors. However, further investigations are needed concerning: i) Incremental Learning (IL), where trained models adapt to learn new postures and control the forgetting of learned postures; ii) MSDs assessment with recognized postures. This study proposed an incremental Convolutional Long Short-Term Memory (CLN) model, investigated effective IL strategies, and evaluated MSDs assessment using recognized postures. Tests with nine workers showed the CLN model with "shallow" convolutional layers achieved high recognition performance (F1 Score) under personalized (0.87) and generalized (0.84) modeling. Generalized "shallow" CLN model under "Many-to-One" IL scheme can balance the adaptation (0.73) and forgetting of learnt subjects (0.74). MSDs assessment using postures recognized from incremental CLN model had minor difference with ground-truth, which demonstrates the high potential for automated MSDs monitoring in construction.

## 1. Introduction

The research is part of a project directed at developing a Data-Driven approach for mitigating the risk of developing awkward posture-related Musculoskeletal Disorders (MSDs), such as the chronic backache and over-exertion, among construction workers. Construction-related MSDs account for 30% of workplace injuries in the U.S. [1] During 2012-2014, employers paid as much as $53.1 billion dollars annually on direct cost for MSDs treatment [2]. Timely detection of awkward postures for ergonomics assessment is becoming a proactive MSDs risk monitoring and prevention strategy [3]. Successful implementation of such strategy requires effective posture recognition from workers and reliable MSDs risk assessment based on recognized postures.

In previous efforts, the authors investigated the potential of a Deep Neural Networks (DNN)-based approach in awkward posture recognition from motion data captured by wearable Inertial Measurement Units (IMUs) [4]. These efforts were built on other successful deployments of Machine Learning (ML)-based models for recognizing construction workers' postures and activities from IMUs output [5-10]. Notably, conventional ML-based models rely on heuristic feature engineering, which can involve engineering bias and ignore the sequential patterns within motion data [11]. Our previous work [4] proposed a seven-layer Convolutional Long Short-Term Memory (CLN)-based DNN architecture, which outperformed benchmark ML models when being tested for recognizing workers' postures during daily tasks. The preliminary results suggest there is a potential to leverage DNN-based automated feature learning and posture modelling to address the limitations of conventional ML-based recognition models.

There is a need to further enhance and validate the use of DNN-based posture recognition models for MSDs risk assessment in construction. Current ML-based recognition models were developed from the full posture datasets of workers' in related studies [4-10, 12]. Full datasets were given prior to model training–this assumed the data and underlying structure are static [13, 14]. However, such assumption may not hold in practice because: i) same postures can be performed differently among people (and even for the same person) over time; ii) trained models may need to learn new posture classes when new training datasets are available; and iii) the low-cost IMUs output can suffer from noise and drift over time [14]. These bring the challenge of "domain variation", where a model trained for *source* domain (old tasks) can fail to achieve high-performance on the *target* domain (new tasks).

Incoming worker subjects with new posture data is the key domain variation in this research context. An ideal recognition model should not only learn postures from *source* subjects but also easily adapt to *target* subjects' postures, without naively model re-training from scratch [14]. Moreover, the updated model should also retain the memory of learned postures from *source* subjects, which allows the recognition model to learn continuously without forgetting. Controlling the forgetting becomes a need when using the updated model for previous tasks, particularly when the previous training data are unavailable [14]. These requirements warrant a recognition model with Incremental Learning (IL) capability in real-world applications [15]. In addition, awkward posture detection is the first step in proactive MSDs prevention. Comparing to rich studies exploring high-performance ML-based recognition models, there is still a need for validating the injury risk assessment using the recognized postures [7] given the presence of recognition errors.

The DNN-based model can continuously update with streaming data coming in batch [16], allowing it to learn new postures from *target* subjects incrementally. It is, nevertheless, important to note the factors influencing the DNN model's performance under IL. Firstly, the model complexity (i.e., depth of DNN architecture) affects its learning capacity and training process [17]. A shallow model is restricted in learning capacity; whereas the deep model's adaptability can be limited by the high complexity and difficulty of convergence in training. Secondly, Learning Rate (LR) controls the speed of adapting a DNN model to *target*. A quick model adaptation allows learning new tasks rapidly, which also brings the risks of impairing the consolidated memory of learned tasks, thus resulting in the forgetting [18]. Additionally, the nature of *source* data can affect the effective IL [19]. In the context of this research, the *source* data can be eighter a single subject or group of subjects. In this case, two learning schemes, namely One-to-One (OtO) and Many-to-One (MtO), are applicable. The OtO approach may adapt a DNN model to a *target* subject with ease. The MtO approach, in comparison, can be excel in capturing subject-invariant features for IL. The inappropriate IL implementation concerning these factors can lead to either the "Catastrophic Forgetting"—an updated model forgets to perform a learned task [20]—or the failure in learning new tasks. It is, therefore, necessary to investigate the proper strategies to implement an incremental posture recognition model in this research context.

The objectives of this paper are two-fold: investigating the feasibility and strategies of applying incremental DNN model for posture recognition from wearable IMU; and evaluating the validity of applying ergonomics rules with recognized postures for MSDs risk assessment in construction. Built on the proposed CLN-based recognition model in previous work [4], we i) investigated the proper CLN architecture for posture recognition; ii) explored the effective IL strategies considering model complexity, LR, and learning schemes; iii) evaluated the validity of using ergonomics assessment rules with recognized postures. The remainder of the paper is organized as follows. Section 2 reviews the closely related studies leveraging ML and wearable IMUs techniques for workers' posture recognition, DNN technique for posture detection under IL, and ergonomics assessment rules for MSDs. Section 3 describes the development of incremental CLN models under different IL strategies. Section 4 presents the evaluation of applying incremental DNN models for risk assessment. Section 5 discusses the results, followed by conclusions in Section 6. Section 7 summarises the limitations in this study and associated further works.

## 2. Research Background

### 2.1 Workers' Posture Detection with Wearable Sensing and Machine Learning

MSDs risk assessment warrants effective monitoring for workers' postures. Conventional observation-based MSDs risk assessment strategies are impractical on construction sites. The complexities of the rapidly changing working conditions will leave safety inspectors overwhelmed [21]. Wearable IMUs sensors emerge as an effective motion-sensing tool in construction [22]. This paper focuses on capturing workers' awkward postures associated with MSDs risks. Posture recognition from IMUs output is usually formulated as a classification problem. Data-driven ML models are gaining increasing research interest for such classification tasks. Table 1 provides a review of related studies leveraging ML techniques and wearable IMUs for motion detection among construction workers, where the ML-based recognition models have shown relatively high accuracy in experiments. This notwithstanding, further works are still needed for both improving recognition performance and applying the models on real jobsites.

Table 1 Review of Related Studies Applying ML-based Recognition Models for IMUs Output

| Models* | Motion Data Collection | | Placement (Numbers) | Recognition Performance | | Safety Risk Assessment |
|---|---|---|---|---|---|---|
| | Subjects | Activities | | Classes | Accuracy | |
| NN [23] | 2 workers | Prescribed activities in experiments | Arm (1) | 3 | around 90% | N/A |
| SVM [5] | 21 workers | Prescribed masonry tasks in experiments | Full-body (17) | 2 | around 91% | N/A |
| SVM [6] | 4 students | Prescribed awkward postures in experiments | Full-body (17) | 9 | around 60-80% | N/A |
| SVM [8] | 10 workers | Prescribed masonry tasks in experiments | Wrist (1) | 4 | around 88% | N/A |
| SVM [10] | 1 student | Prescribed award postures in experiments | Full-body (5) | 7 | around 74-83% | N/A |
| SVM [9] | 25 students | Prescribed activities in experiments | Leg & wrist (2) | 8 | 89% | N/A |
| SVM [7] | 2 workers | prescribed activities in experiments. | Arm & wrist (2) | 3 | up to 90.2% | OSHA Rules |
| LSTM [24] | 3 students | Prescribed activities in experiments | Hip & neck (2) | 11 | up to 94.7% | N/A |
| CLN [4] | 4 workers | Natural postures in daily tasks | Full-body (5) | 8 | 0.85(Macro F1 Score) | N/A |

*The recognition models were those achieving highest recognition performance in the tests of corresponding studies. NN-Neural Networks, SVM-Support Vector Machine, LSTM-Long Short-Term Memory.

Conventional ML models developed in related studies [5-10, 23] typically adopted a "sliding-window-based analysis pipeline" [11]. The manual heuristic feature engineering renders a biased process [11], hindering effective motion feature construction [25]. Conventional ML models also lack the mechanism of capturing temporal patterns within motion data streams, thus treating the translational and temporal motion data as static [12, 25]. Additionally, feature engineering and model parameter tuning are conducted independently, without optimizing these intervening processing together when training recognition models [11]. These unsolved problems can result in a sub-optimal posture recognition model.

DNN is the ML technique that uses representation learning to discover features in raw data [26], which automates feature engineering with minimal human efforts [11]. Recent studies have started to explore leveraging DNN models' advantages in automated feature engineering for high-performance workers' motion detection from IMUs. Kim and Cho [24] achieved a high recognition accuracy (94.7%) by using the Long Short-Term Memory (LSTM)-based DNN model, which leveraged discriminative temporal motion patterns for workers' activity recognition. The authors also proposed the use of a seven-layer CLN-based model [4], which showed the feasibility of integrating automated feature engineering and sequential pattern learning to improve the recognition performance of conventional ML models. However, developing DL-based models for processing motion sensing data is still an open area. Further investigation is needed regarding the proper deep model architectures for high-performance motion detection [11].

Challenges also emerge when applying the recognition models on real jobsite scenarios. Motion data used for developing and testing recognition models were collected in controlled experiments in related studies. Prescribed activities or postures were conducted by real workers [5, 7, 8, 23] or students [6, 9, 10, 24], imitating the real construction activities. Recognition models were developed based on the given full motion dataset, which assumed the motion data and underlaying structure are static [13, 14]. Such an assumption may not hold considering both inter-subject and intra-subject variability. The same postures can vary among workers executing similar tasks in workplace. The same posture can be performed differently for even the same worker doing a routine task over time. Furthermore, the recognition model developed for the existing group of workers may not be applicable to a new worker, particularly when new posture classes emerge. Additionally, low-cost IMUs sensors may suffer from noise and drift over time. The classifiers should be able to adapt to changes in the sensor output (even changes of sensors). How recognition models can adapt to these dynamic variations and still retain high performance brings an implementation challenge. It is, therefore, necessary to investigate proper feature engineering, model optimization, and model adaptation strategies for applying Data-Driven posture recognition models in real-life scenarios.

Additionally, despite the vibrant research in enhancing the performance of recognition models, few studies have started to evaluate the validity of workers' safety assessment using the output from developed recognition models. Nath, Chaspari and Behzadan [7]'s work investigated the validity of assessing the repetition of workers' activities (detected from ML models) under Occupational Safety and Health Administration (OSHA)'s regulations. Given errors made by recognition models, further validation is needed when applying posture-based ergonomics rules with recognized postures.

## 2.2 Deep Neural Networks-based Posture Recognition Models

### 2.2.1 Automated Feature Learning and Model Development

DNN models have achieved state-of-the-art performance for pattern recognition tasks with images and video streams [11]. The synchronized multi-channel motion data resemble a 2D "Image" [25], enabling the application of DNN models on IMUs output. Hybrid DNN models, integrating the strength of different functional layers, tend to dominate the landscape of this research area [25].

The multi-layer Convolutional Neural Networks (CNN) model automatically extracts rich features with increasing complexity from input data, eliminating the tedious manual feature engineering. The LSTM model, an Recurrent Neural Networks (RNN) derivate, manages to learn long-term sequential patterns from input data. On-going studies have shown the great potential of using CNN with LSTM for processing sensor output [11], such as recognizing daily living activities [12, 19, 27] and monitoring sleep condition [28].

It is worth noting that, unlike well-developed DNN architectures (e.g., Inception V3, VGG-16, and ResNet-50) for image and video processing, there is a lack of pre-trained and ready-to-use DNN architectures for different application scenarios with multi-channel Wearable Sensors (WS) [11]. The authors' initial study evaluated a seven-layer CLN architecture on a relatively small dataset (of four workers) [4]. Little is known regarding how the CLN architecture can be optimized for improving both recognition accuracy and computational efficiency. A larger motion dataset incorporating more posture data allows investigation and validation of the proper CLN architecture. Built on the previous study, we expanded the dataset and further investigated a high-performance model architecture for posture recognition.

### 2.2.2 Incremental Learning for Posture Recognition Models

Posture recognition models trained once using short-term datasets may not be reliable for long-term applications under dynamic conditions. The dynamics with WS-based posture detection makes the native solution, repetitive model re-training, both time-consuming and impractical. Recognition models are, therefore, expected to i) learn novel information from incoming training datasets; ii) add new classes for classification; and iii) update without previous training dataset [14], given the burden of data storage. As a result, Transfer Learning (TL) is warranted [15].

TL is the ability to extend what has been learned from one *source* domain to another non-identical but similar *target* domain sharing common features [29]. The major domain variation in WS-based posture recognition can be attributed to user difference [15], which can bring both sensor changes (e.g., placement variation) and posture changes (inter-subject posture variation). The TL can be implemented via IL, where the trained recognition model continuously adapts to constantly arriving data stream [13]. The incremental adaptation eliminates the model re-training from scratch and user-interruption [30]. However, conventional ML models may not be used directly under IL. One of such non-incremental models is SVM, a widely used classification model for recognizing workers' postures (see Table 1). SVM models need to be re-trained from scratch when new training data are available. An SVM model trained on two different tasks in sequence will completely forget the first task [31]. Conventional ML models require sophisticated adaptions for IL, e.g., ensemble [32] or Prototype-based methods [13].

DNN models can work in the regime of IL, where they adapt to new classes and tasks [16]. Training DNN-based recognition models often includes a supervised model fine-tuning. The weights in DNN models are usually optimized using the Stochastic Gradient Descent (SDG) algorithm. SGD estimates the error gradient for the current model state using examples from the training dataset; it then updates the model's weights using the backpropagation of errors. The amount that the weights are updated during training is referred to as step size or "Learning Rate (LR)". The DNN model with optimized weights from a supervised fine-tuning process renders a parametrized learner [16]. DNN models randomly initialize the weights when training from scratch, followed by the weight optimization for learning salient and discriminative features corresponding to different classes for classification. When applying a pre-trained DNN model with learned weights, the newly collected (labeled) data will be used for updating saved weights via backpropagation. The incremental training with new data becomes a straightforward and easy-to-use IL approach for DNN models [17].

Notably, continuous DNN model training under IL can encounter the "stability-plasticity" dilemma [17]. The quick model updating with new data allows rapid adaption to new tasks, while the model can forget old tasks equally quickly at the same time. Similarly, the memory of old tasks is preserved longer, but model reactivity decreases in the case of slow model adaption. This is a well-known constraint for artificial as well as biological learning systems [33]. The failure to address such a dilemma can result in the "Catastrophic Forgetting" [20], given the DNN models' high adaptability [31]. The forgetting effect in this study reveals as a trained DNN model forgets the learned postures after learning a new subject's posture. Therefore, both the adaptability to new tasks and forgetting effect on learnt tasks should be considered when evaluating IL performance.

The IL performance of DNN-based models can be affected by model complexity, LR, and learning scheme. The model performance converges slowly under an overly complex deep model, whereas a simple and shallow model architecture limits its learning capacity [17]. LR may be the most important hyperparameter for fine-tuning the DNN models [34]. Choosing the LR is also challenging as a value too small may result in a long training process that could get stuck, whereas a value too large may result in learning a sub-optimal set of weights too fast or an unstable training process. Additionally, the trained recognition models can be developed from either one subject (OtO scheme) or a group of subjects (MtO scheme) before adapting to the new subject. The OtO scheme may adapt a DNN model to a new subject with ease, given less memory of learnt tasks need to be retained. The MtO scheme may be better at adapting to new subject's postures and ameliorating the forgetting, which benefits from the DNN model developed to extract subject-invariant features. Proper DNN model complexity, LR setup, and learning scheme can potentially balance the model's performance in both adaptation and forgetting. Therefore, there is a need for systematic investigation concerning i) the feasibility of implementing the incremental DNN models; and ii) how the model architecture and training strategy can influence the model's performance under IL.

### 2.3 Posture-based Ergonomics Assessment for Proactive MSDs Prevention

Epidemiological studies have established that physical factors, such as construction-related awkward working postures, pose the high risks for MSDs [35]. Consequently, there have been efforts on using observation-based ergonomics rules to assess the level of exposure to awkward postures in the workplace. Common ergonomic rules include "Rapid Upper Limb Assessment" (RULA) [36] and "Rapid Entire Body Assessment" (REBA) [37] for postures analysis through body joint angles; Ovako Working Posture Analysing System (OWAS) [38] and its extension [39] for evaluating awkward postures' riskiness by measuring posture usage time; the ISO 11226:2000 [40], which assesses postures considering both level of awkwardness and holding time; and OSHA's ergonomics rules for assessing MSDs risk levels using working posture duration and frequency [7]. Additionally, Miedema, Douwes and Dul [41] also provide the threshold for awkward postures' Maximum Holding Time (MHT) based on experimental results.

The ergonomics rules are widely applied with automated safety monitoring technologies [7, 42-44]. These rules serve as the criteria for evaluating the workers' injury risk levels based on the captured safety information. In particular, the OWAS provides actionable recommendations for correcting awkward postures at a varying level of urgency, considering the proportion of posture usage. The MHT criteria allow identifying postures held for too long, thus triggering alarms in real-time [42]. Both OWAS and MHT criteria provide guidance for defining approximate human postures, allowing one to use recognized postures for injury risk assessment. Notably, posture misclassifications from recognition models may lead to false alarm or injury risk underestimation. It is, therefore, important to evaluate the validity of using recognized postures with ergonomics rules. The evaluation helps with understanding how recognition errors can influence risk assessment results. This paper evaluated the ergonomics assessment results using both OWAS and MHT rules, based on the recognized postures from proposed incremental models.

## 3. Development and Implementation of Incremental Convolutional LSTM Model

This section describes the development of proposed incremental CLN model (Section 3.1) and IL strategies implemented for the proposed model (Section 3.2).

### 3.1 Convolutional LSTM Model Architecture

Figure 1 presents the schematic diagram of the proposed incremental CLN model architecture. The following sections describe the model design regarding each component.

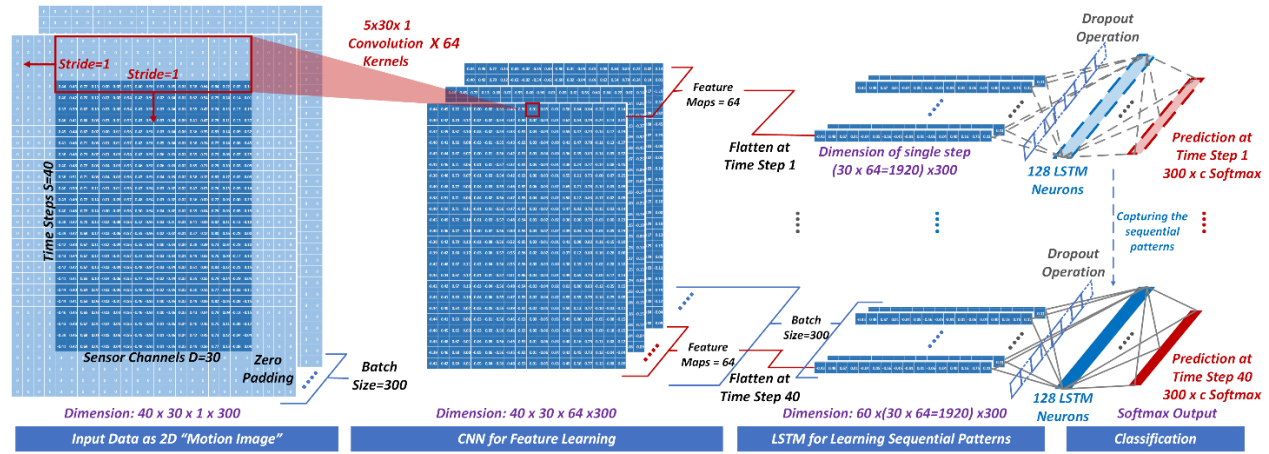

Figure 1. CLN Conceptual architecture integrating one-layer CNN and one-layer LSTM. The parameter setup is discussed in Section 3.1.4

### 3.1.1. Constructing "Motion Image" from IMUs Output

Converting multi-channel time-series motion data from IMUs into 2D "Motion Image" description enables learning discriminative features [45]. The continuous motion data were segmented into consecutive equal-size windows (window size is discussed in Section 4.2.1). Each window became an "image", where the "pixel value" was the sensor output from a specific channel ($d_i$) at a certain time step ($s_i$). The channel-wise normalization was applied to each window to addresses the unit difference across channels, which was achieved by centering to the mean and scaled to unit variance. Combining normalized channels in the same layer resulted in a "Motion Image", which had the dimension of "S (time steps in a window) by D (channels) by one-layer depth".

### 3.1.2. Convolutional and LSTM Layers for Automated Feature Learning

Convolutional layers conduct the convolution operation between the input "Motion Image" and convolutional kernels. The kernels optimized under the supervised learning process attempt to maximize their activation level for data subsets in the same class. The optimized kernel weight serves as a feature detector. Combining discriminative features learned by kernels renders a feature map. Such feature map identifies a specific salient pattern of targets (e.g., postures with corresponding motion data patterns). Stacked convolutional layers are becoming the "de facto" approach for automated feature learning [12], where deeper layers progressively represent the prior layer's output in a more abstract way and discover highly discriminative features via the hierarchical representation of motion data. Following the last convolutional layer, the Flatten operation built a fully-connected dense layer, thus converting feature maps for a window into a 1D vector. The flattened 1D vector contains the learnt features from convolutional layers for charactering the posture corresponding to a window.

### 3.1.3. LSTM Layer for Sequential Patterns Learning

The LSTM extends the conventional Recurrent Neural Networks' abilities in learning long-term temporary relationships [12]. Figure 2 shows the LSTM working procedure following Olah [46]'s work. The LSTM differentiates long-term memory ($c_t$) and short-term memory ($h_t$), then uses the "gate" to handle historical information in a more intelligent way [47]. LSTM firstly learns which information should be kept or forgotten in the long-term memory $c_t$ by a forget gate $f_t$ (Eq. 1). Then it calculates the candidate new information $\tilde{c}_t$ (Eq. 3) to be added into $c_t$. LSTM saves useful information from the current input $x_t$ and stores it in the $c_t$ with learnt input gate $i_t$ (Eq. 2). Next, $c_t$ is updated using $f_t, i_t, \tilde{c}_t$ and previous cell state $c_{t-1}$ (Eq. 4). Finally, the model determines which part of $c_t$ should be focused on for the current work $h_t$. The output gate $o_t$ learnt from Eq. 5 is applied to update $h_t$ in Eq. 6. In Eq. 1 to Eq. 6, $\sigma$ is the non-linear activation function; $W_{ij}$ is the weight matrix describing from-to relationships (e.g., $W_{ho}$ denotes the hidden-output gate).

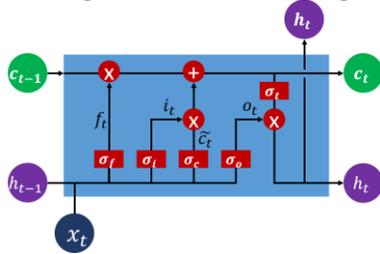

$$f_t = \sigma_f(W_{xf}x_t + W_{hf}h_{t-1} + b_f) \quad \text{Eq. 1}$$
$$i_t = \sigma_i(W_{xi}x_t + W_{hi}h_{t-1} + b_i) \quad \text{Eq. 2}$$
$$\tilde{c}_t = \sigma_c(W_{xc}x_t + W_{hc}h_{t-1} + b_c) \quad \text{Eq. 3}$$
$$c_t = f_t c_{t-1} + i_t \tilde{c}_t \quad \text{Eq. 4}$$
$$o_t = \sigma_o(W_{xo}x_t + W_{ho}h_{t-1} + b_o) \quad \text{Eq. 5}$$
$$h_t = o_t \sigma_t(c_t) \quad \text{Eq. 6}$$

Figure 2. LSTM Model Working Process

As shown in Figure 1, LSTM layers was connected with the flattened feature maps for learning sequential patterns in a window. Flattening all feature maps within one "Motion Image" ignores temporal dependencies along with the time step. LSTM addresses this problem by flattening feature maps only along the depth dimension, thus preserving the time step dimension for capturing sequential patterns. The 50%-dropout layer controls model overfitting by randomly setting the activation of half of the units in a subsequent layer as zero. The softmax layer fully collected with LSTM neurons can yield a class probability distribution of samples in the batch. Each sample was classified by the class label with the highest probability. Notably, the activation information in LSTM neurons at each time step is passed on to the next. The more time steps LSTM neurons have "processed", the more informative the model will be [12]. Therefore, prediction at the last time step was used as the recognition result, after the full sequence within a window was processed.

### 3.1.4. Convolutional LSTM Model Setup

Convolutional layer depth is a key hyperparameter influencing DNN-based model complexity and performance [4, 12]. We, therefore, investigated the optimal CLN model architecture by varying the convolutional layer depth ranging from one and five, as suggested in [12]. We adopted the same parameter setup for each convolutional layer, which has shown promising results in the initial study [4]. Specifically, each convolutional layer had 64 kernels with a size of 5 by 30, stride of 1×1, and zero-padding. The two-layer LSTM architecture with 128 neurons per layer was applied as recommended in [12, 48]. The model was expressed as $C(64) \times N - RL(128) \times 2 - Sm$ (or C$N$L2 for short) as suggested in [49], where $C$, $RL$, and $Sm$ were CNN, LSTM, and softmax layers, respectively. The hyperbolic tangent function (tanh) was chose as activation. The entire dataset for training CLN models was divided into multiple (non-overlapping) batches with a size of 300 windows/batch. The batches were fed into the model one by one for effective model training [16, 34].

CLN models were trained under Supervised Learning. We adopted Adam optimizer, which is recommended for training DNN models with convolutional layers [50]. Adam extends the classical SGD by leveraging an adaptive LR (approximately bounded by the initial LR, i.e. $10^{-3}$ by default without decay) based on the average of recent magnitudes of gradients for the weight (i.e., how quickly the LR changes). Adaptive learning speed can potentially regularize the deep model and ameliorating the forgetting effect under IL [18]. We, therefore, used default LR ($10^{-3}$) for developing and testing non-incremental CLN models and adjusted levels of LR (LR1-$10^{-2}$, LR2-$10^{-3}$, and LR3-$10^{-4}$ in Adam optimizer) when evaluating the incremental CLN models.

### 3.2 Incremental Learning Scheme

The *source* dataset under the IL context of this study is postures of the existing worker subject, while the *target* dataset is postures of an incoming new subject. The IL scheme describes how a recognition model is initially developed from *source* and continually adapted to the *target*. Both personalized and generalized modeling can be applied to develop the *source* CLN model from scratch. Personalized model (Figure 3-a) emphasizes learning subject-specific features for high recognition performance, whereas it requires repetitive model re-training for each new subject. Generalized model (Figure 3-b), trained to learn subject-invariant features, aims at recognizing multiple subjects' postures using one generic model. However, the generalized model may not achieve reliable performance when being applied for a new subject. One ideal situation is "adaptive personalization" [11], which adapts the trained model to a new subject as a personalized model. The adaptive personalization aligns with goal of IL.

We started by identifying the proper CLN model architectures (in terms of convolutional layer depth) under both personalized and generalized modeling. The personalized and generalized models' IL performance was then evaluated when adapting to the *target* subject under both One-to-One (OtO) and Many-to-One (MtO) strategies, respectively. By IL performance, we evaluated two aspects concerning: i) incremental performance, which denotes the adaptability to the *target* subject's postures when using the incremental model already trained for *source* subject(s); ii) forgetting performance, which denotes the incremental model's performance on learnt subject(s)' postures after being adapted to the target subject. Two IL schemes investigated are described in detail below.

**One-to-One**. In Figure 3-c, the personalized model $M_s$ developed from the *source* subject $S_s$ continuously adapted to the *target* subject $S_t$. The incremental model $M_{s \to t}$ was trained via a personalized modelling approach on $S_t$; however, the difference was that $M_s$ was re-loaded as a starting point for training $M_{s \to t}$, instead of re-training from scratch on $S_t$. Model $M_t$ trained solely for $S_t$ denotes a personalised model on *target*. In this case, the recognition performances of $M_{s \to t}$ tested on $S_t$ and $S_s$ represent incremental and forgetting performances, respectively. The performance difference between $M_{s \to t}$ and baseline model $M_t$ on $S_t$ reflects the model's adaptability, where a small difference represents a higher adaptability. The performance difference between $M_{s \to t}$ and baseline model $M_s$ on $S_s$ reflects the forgetting effect under IL, where a small difference represents better control of forgetting.

**Many-to-One**. In Figure 3-d, $S_s$ became a group of subjects while $S_t$ was still one subject. The "leave-one-out" was applied to evaluate the MtO-based incremental performance on *target* subject, where the rest subjects were used as *source* for training $M_s$ from scratch. Similarly, $M_{s \to t}$ was evaluated on $S_t$ and $S_s$ to test the incremental and forgetting performances, then compared with $M_t$ and $M_s$ to evaluate the model adaptability and forgetting effect, respectively.

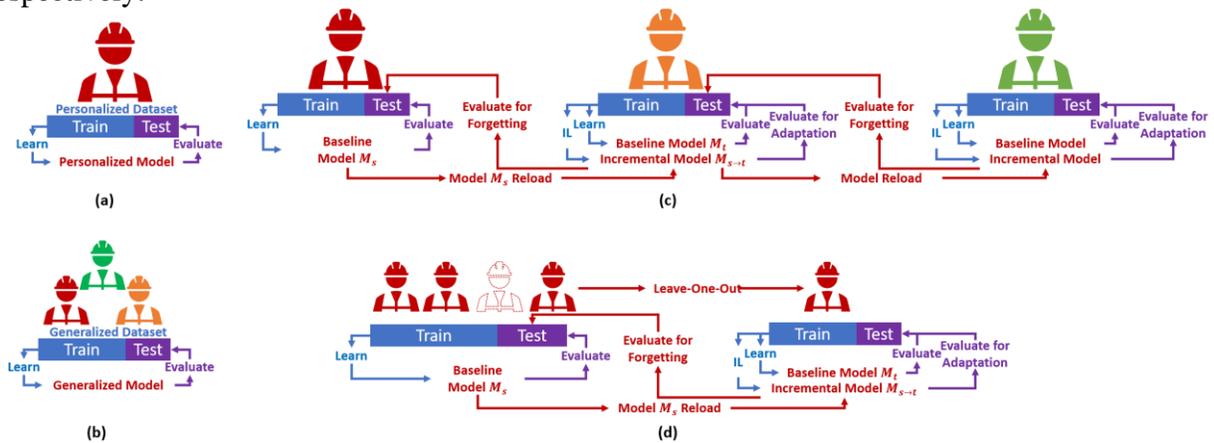

Figure 3. Model Development Approaches. Non-incremental Learning: (a) Personalized Modelling; (b) Generalized Modelling; Incremental Learning: (c) OtO Scheme; (d) MtO Scheme

## 4 Evaluation of Incremental Posture Recognition Models

This section reports how the proposed incremental posture recognition models were developed and evaluated using workers' posture data collected on construction sites. It then describes the validation of ergonomics assessment using postures detected from developed recognition models.

### 4.1 Motion Data Collection on Construction Site

Nine worker subjects from five different trades were recruited from a residential building construction project. The average work experience of subjects is 14.5 years in current trade. Workers' consent was obtained following Institutional Review Board (IRB) approved protocols. Five IMUs sensors (Meta Motion C [51]) were deployed at the forehead (on the front of hardhat), chest center, right upper arm, right thigh, and right calf (Figure 4) for motion data collection, considering all subjects were right-handed. The sensor placements were selected according to the human body segments and landmarks suggested in [52]. Each IMUs sensor captured motion data from 6 channels (tri-axial channels for accelerometer and gyroscope). The 30-channel motion data were collected from all sensor placements. Subjects performed their routine tasks for 20 to 30 minutes with their naturalistic postures. Workers' postures were videotaped as ground truth reference.

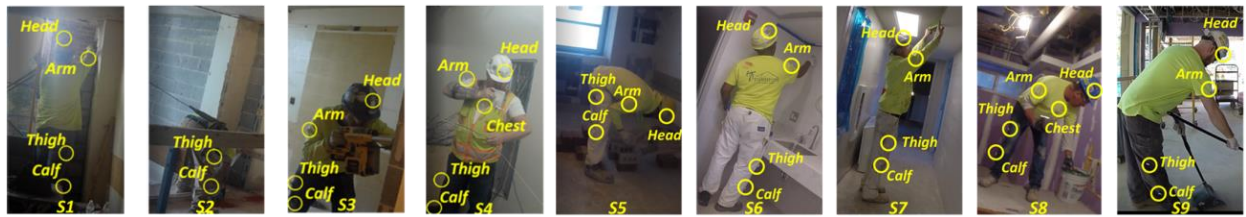

Figure 4. Subjects Working with Sensors (the sensors blocked are not circled)

Table 2. Description of Collect Motion Dataset

| Subjects Tasks | Motion Dataset | Postures (Percentage-%) | | | | | | | | | Posture Label Explanation |
|---|---|---|---|---|---|---|---|---|---|---|---|
| | | BT | KN | LB | MO | TR | SQ | ST | WK | WO | |
| S1 Masonry *Bricklaying* | 30.27min @20hz@30 channels | 14.7 | 2.0 | 12.3 | 0.0 | 0.0 | 0.0 | 52.4 | 3.4 | 7.2 | BT-Static bending, minor movement with bending, minor literal bending, and short-term pick up. |
| S2 Labour *Guardrail Installation* | 30.27min @40hz@24 channels* | 72.9 | 0.0 | 0.0 | 0.0 | 4.3 | 0.0 | 12.5 | 9.1 | 0.0 | |
| S3 Electrician *Conduit Installation* | 18.50min @40hz@30 channels | 13.6 | 46.7 | 0.0 | 0.0 | 15.0 | 3.0 | 22.0 | 0.0 | 0.0 | KN-Kneel on one leg and both legs. LB-Literal bend MO-climbing ladders. SQ- Squatting. ST- Standing with minor movement. WK-Walk. TR-Transitional postures between consecutive postures. WO- Overhead work with at least one arm. |
| S4 Electrician *Wire Pulling* | 18.50min @40hz@30 channels | 12.3 | 0.0 | 0.0 | 0.0 | 0.0 | 0.0 | 71.5 | 12.3 | 0.0 | |
| S5 Labour *Cleaning Work* | 19.38min @40hz@30 channels | 18.7 | 0.0 | 0.0 | 1.8 | 0.0 | 0.0 | 23.6 | 19.9 | 32.6 | |
| S6 Painter *Wall Painting* | 19.63min @40hz@30 channels | 10.5 | 0.0 | 0.0 | 0.0 | 0.0 | 0.0 | 17.0 | 14.1 | 54.4 | |
| S7 Painter *Stick Tapes* | 20.47min @40hz@30 channels | 1.7 | 0.0 | 0.0 | 0.0 | 0.0 | 0.0 | 5.9 | 2.1 | 83.6 | |
| S8 Carpenter *Wall Plastering* | 12.50min @40hz@30 channels | 7.1 | 0.0 | 0.0 | 1.8 | 0.0 | 0.0 | 27.3 | 8.0 | 23.1 | |
| S9 Labour *Cleaning* | 18.53min @40hz@30 channels | 10.0 | 0.0 | 0.0 | 0.0 | 0.0 | 0.0 | 26.5 | 56.4 | 0.5 | |

*The six channels from arm sensor was not considered due to sensor malfunction.

## 4.2 Data Preparation

### 4.2.1 Data Segmentation and Pre-processing

The window size of 0.5-2.5 s is a commonly used range for daily-living activity recognition [53]. Our previous work found the window of 1.0-1.3 s can achieve high posture recognition performance [54]. We adopted a 1.0 s window for segmenting the streaming motion data. The sensor frequency was set as 50 Hz for S2 to S9. 40 Hz was used as a cut-off for each channel to remove lower frequency windows. A common subset of remaining windows across all channels was then combined. Finally, 40 data records were randomly sampled with preserved sequences from each window across all channels. The downsampling resulted in a 40 Hz motion dataset for subjects S2 to S9. The data for S1 were collected at 25 Hz and, therefore, downsampled to 20 Hz.

### 4.2.2 Posture Labelling

Each data record (representing the sensor output from all channels at time step $t$) was labeled with video reference after pre-processing. The workers' postures were labeled by referencing the posture definition in OWAS. Table 2 describes the distribution of labels and associated postures. Motion data without video references (due to block of sight) were not considered. The labeled motion dataset was re-segmented with a 50% overlap to capture the posture transitions between consecutive windows. Each window was labeled using the label of majority records it contained.

### 4.3 Model Training Setup

#### 4.3.1 Dataset Splitting

The Stratified Random Shuffle (SRS) in Figure 5 was used for the "train-validation-test" splitting. Stratified sampling keeps the same class distribution in both "train" and "test" datasets. The data shuffling is recommended for effectively training "mini-batch" based DNN models [16], particularly when datasets naturally grouping the same classes in sequence [34], like the posture datasets in this study. In addition, the data shuffling also reduces the potential drift in motion data [55], which can be caused by unstable sensor output from IMUs overtime [56].

For a given dataset in this study, the "train" and "test" subsets were split as 9:1. The "train" dataset was further split into "training" and "validation" datasets using a ratio of 8:2. The splitting ratios were set to preserve more data for training the complex DNN-based models. The "train-test" split was repeated five rounds under SRS to reduce the bias in dataset splitting. The "training-validation" split was performed for once to shorten the model training time. The same training process yields a DNN model with slightly different performance on the same test dataset, which is caused by the random parameter initialization and stochastic optimization algorithms. We performed repetitive model training and testing under different splitting to reduce the bias in model performance evaluation.

#### 4.3.2 Recognition Performance Evaluation Metric

Workers' natural posture distribution can be highly unbalanced between classes. Macro F1 Score was adopted to account for such imbalance. F1-socre was calculated by the harmonic mean of Precision (Eq. 7) and Recall (Eq. 8). The average F1-score was used as evaluation metric after acquiring the F1-score for each class. Table 2 shows the awkward posture can be either majority or minority. It is, therefore, appropriate to train recognition models for achieving balanced performance for both majority and minority postures, resulting in a Macro F1 Score (Eq. 9) where $N$ denotes the number of label classes. Higher Macro F1 Score denotes a higher classification performance on unbalanced dataset.

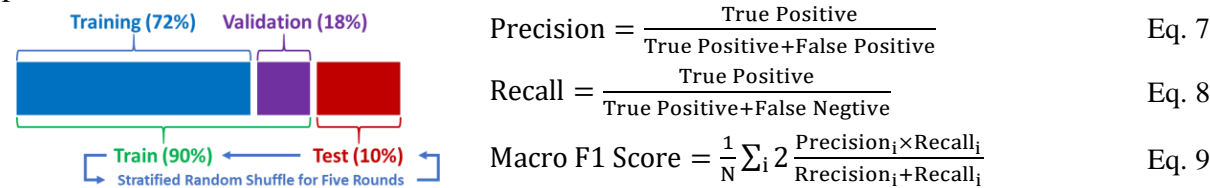

$$\text{Precision} = \frac{\text{True Positive}}{\text{True Positive} + \text{False Positive}} \quad \text{Eq. 7}$$

$$\text{Recall} = \frac{\text{True Positive}}{\text{True Positive} + \text{False Negtive}} \quad \text{Eq. 8}$$

$$\text{Macro F1 Score} = \frac{1}{N}\sum_i 2\frac{\text{Precision}_i \times \text{Recall}_i}{\text{Rrecision}_i + \text{Recall}_i} \quad \text{Eq. 9}$$

Figure 5. Data Distribution under SRS

#### 4.3.3 Model Training Checkpoint

When training DNN models, one epoch represents all batches of training data that have passed both forward and backward through the model for once. Multiple epochs are typically applied to fully train the DNN models with limited data. The model performance may not increase consistently after every epoch during training. We trained the DNN models until their performance ceased to increase when being tested on validation datasets. The model training checkpoint was set to only save the trained model with improved performance (Macro F1 Score) in an "overwritten" way. We set the total epoch as 300 to fully train the model based on the observation that model training performance became stable around 100-150 epochs (see Figure 7-c). Therefore, DNN models with the highest performance after 300 epochs were saved.

All DNN models were developed using Keras 2.2.2 [57] (TensorFlow 1.9.0 GPU version). The models were all implemented on a Windows 10 PC (Intel Core i7-7700 CPU@ 2.8 GHz, 16GB RAM, NIVIDA GeForce GTX 1060 GPU@16GB RAM). The code is available at [58].

### 4.4 Posture Recognition Model Implementation and Evaluation

#### 4.4.1 Personalized and Generalized Modeling

We identified the proper CLN architectures by comparing model performance with varying convolutional layers under both personalized and generalized modelling. Each pre-processed dataset of S1 to S9 was used for personalized modeling. The generalized dataset was constructed by combining the datasets of S3-S9. Two subjects were excluded due to the low frequency of motion data (S1) and lack of arm sensor output (S2). The TR posture was deleted from S3 when constructing the generalized datasets. Results are discussed in Section 5.1.

#### 4.4.2 Incremental Modeling

The OtO scheme was iteratively conducted from S3 to S9. The personalized CLN model initially trained from scratch for S3 and continuously adapted to the next subject in sequence. Under the MtO scheme, the "leave-one-out" was repeated for each subject in the generalized dataset (combining S3-S9) to evaluate the performance of incremental CLN model. Three levels of LR (LR1-$10^{-2}$, LR2-$10^{-3}$, and LR3-$10^{-4}$) for the CLN model with varying convolutional layer depth were tested under both OtO and MtO schemes. Section 5 describes the systematic evaluation of the incremental CLN model under different IL strategies regarding model complexity, LR, and learning schemes.

#### 4.4.3 Evaluation of Posture Recognition from Incremental Learning Models

After identifying the proper IL strategy for the incremental CLN model, this study examined the model's incremental and forgetting performance on each type of postures. For all postures, we compared the performance difference between incremental and baseline (personalized) CLN models on each subject (S3-S9) as *target*. A smaller performance difference indicates the posture has a higher potential of being incrementally learnt across subjects. Similarly, we also compared the forgetting effect of each posture learnt from *source*. A smaller forgetting effect suggests the posture is more likely to be remembered under IL. The results are reported in Section 5.3.1.

This study further investigated the effective sensor placement and channels for recognizing workers' posture when using the incremental CLN model. All sensor channels were divided as 10 groups with respect to 5 placements (arm, calf, chest, head, and thigh) and 2 sensor units (accelerometer and gyroscope). E.g., "chest_acc" represent the group of sensor channels from accelerometer units placed at the chest. Despite the DNN-based model are typically applied as a "black-box" with less interpretability, the "Feature Permutation" [59] approach can be used to evaluate the importance of each sensor channel groups played in the proposed incremental CLN model. The group of channels is not "influential" if shuffling their values leaves the model performance unchanged, as the model relies less on these channels for postures recognition. In this sense, we iteratively permuted each sensor channel group in test datasets; then compared the change of both incremental and forgetting performances for each posture. The "influential" sensor placements and units regarding each posture can be identified through comparing performance change. The results are summarized in Section 5.3.2.

**4.5 Posture-based Ergonomics Risk Assessment**

**4.5.1 Assessment Rules Implementation**

The OWSA considers different body parts positions when evaluating ergonomics risks. Each body part has an associated threshold concerning the posture proportion in unit working time. We adopted a conservative criterion, which used the strictest threshold among all the affected parts' positions when assessing a specific posture. E.g., "one leg kneeling" consists back bent, both limbs below shoulder, and kneeling. The corresponding thresholds are 30%, 100%, and 20%, respectively. Therefore, 20% is the threshold for posture correction in this case. Three levels of ergonomics risks were determined accordingly for captured awkward postures (BT, KN, SQ, and WO) in this study: I (≤20%)-no action needed, II (20%-50%)-posture corrections soon, and III (50%≤)-correction immediately. The prolonged posture should not be held more than 20% of the MHT as specified in Miedema, Douwes and Dul [41]'s study. We adopted the conservative MHT thresholds by setting: 30 seconds for uncomfortable postures (BT, KN, LB, SQ, and WO); 3 minutes for comfortable postures (WK, MO, and ST). The authors developed an algorithm (pseudo-code in Table A. 1) to implement the above ergonomics rules for MSDs risk assessment.

**4.5.2 Evaluation of MSDs Risk Assessment with Recognized Postures**

The MSDs risk assessment with recognized postures from incremental CLN model was evaluated via a quasi-experiment. The quasi-experiment simulated the scenario where the *target* subject firstly conducted a set of "prescribed" postures in a longer session for adapting the incremental model from *source* to *target* subject; then the *target* subject conducted the same set of postures again, but in a shorter session, for testing the posture recognition and MSDs assessment with recognized postures. To conduct the quasi-experiment, we split the motion data of every continuous posture from *target* subject as two parts: the first 90% was used as "train" dataset which was further randomly split (with shuffling) into "training" and "validation" subsets using the ratio of 4:1; the last 10% was used as "test" dataset (Figure 6).

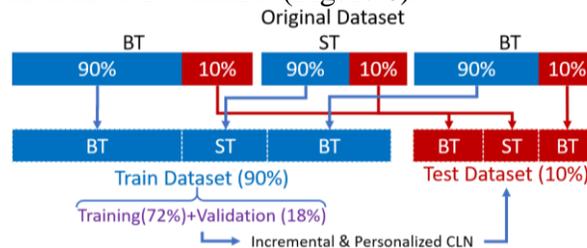

Figure 6. Dataset Splitting for Evaluating Posture-based Ergonomics Assessment Rules

We compared the ergonomics risk assessment results using postures from ground truth and incremental CLN model. The increment model was developed under MtO scheme (considering it outperformed OtO scheme, see Section 5 for detail). The evaluation was repeated for each of S3 to S9 as *target* subject, whereas the rest subjects were used for developing generalized CLN models. The results are discussed in Section 5.4.

## 5 Results and Discussion

This section discusses the test results with respect to i) investigation of a proper CLN architecture for high-performance posture recognition (Section 3.1); ii) exploration of effective IL strategies (Section 5.2); iii) evaluation of applying incremental CLN model for posture recognition (Section 5.3), and iv) evaluation of using recognized postures for ergonomics assessment (Section 3.1).

### 5.1 Identifying the CLN Model with Proper Convolutional Layer Depth

Figure 7-a depicts the evaluation results of the CLN model with convolutional layer depth varying between one and five while preserving the two-layer LSTM. These results suggest that CLN model with "shallow" convolutional layers tended to provide higher recognition performance under both personalized (C2L2) modeling and generalized (C1L2) modeling. Increasing the convolutional layer depth did not improve the recognition performance of CLN model. Particularly, the overly deep architecture (C5L2) gave the lowest model performance under both personalized and generalized CLN models. These might be explained by the greater model depth increases the number of parameters significantly (Figure 7-b). In addition to the greater depth with limited training data being overfitting, the "gradient vanishing" problem can also emerge. In this case, the gradient decreases exponentially in the initial layers after propagating through multiple activation layers, resulting in inefficient model training as the weights in initial layers update slowly. Greater convolutional layer depth also led to increased computational complexity. Both the training time per epoch (Figure 7-c) and posture recognition time during model operation increased (Figure 7-b). Therefore, C2L2 and C1L2 were identified as proper personalized and generalized CLN architecture, respectively, for further IL implementation.

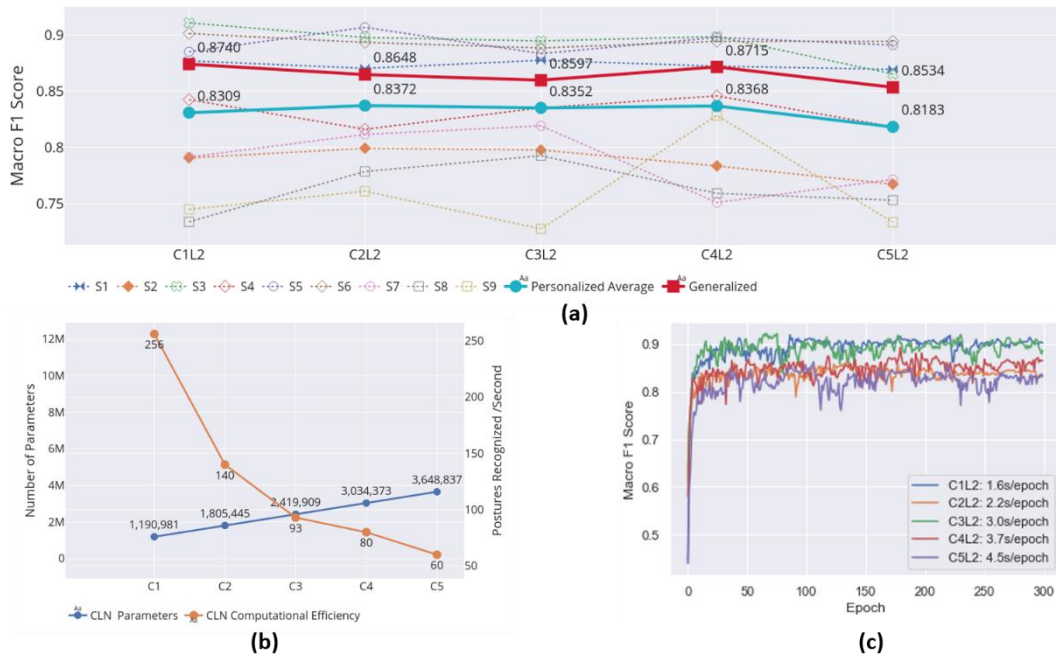

Figure 7. Analysis of Convolutional Layer Depth-Influence on Model Performance. (a) Analysis of CLN model performance. The dots for S1-S9 represent the average performance over five-round SRS. The "Personalized Average" denotes the average personalized performance of all subjects. The "Generalized" represents the average performance of generalized model over five-round SRS. (b) Analysis of CLN model complexity, using the model test result on S3's dataset as an example. (c) CLN Model Training Process (monitoring F1 score on validation datasets). In the legend, "C$N$L2: Xs/epoch" represents training a $C(64) \times N - RL(128) \times 2 - Sm$ model for one epoch requires X seconds.

## 5.2

## 5.2 Evaluation of Incremental Strategies for CLN Models

This section evaluates the performance of incremental CLN models under three levels of LR, two learning schemes, and varying model complexity. Figure 7 shows the "deep" C4L2 model gave a close performance to optimal "shallow" CLN architectures under both personalized and generalized modeling. The authors, therefore, used the C4L2 model as the "deep" CLN architecture when investigating the influence of model complexity on IL. Figure 8 and Figure 11 describe the incremental CLN model's performance (regarding both incremental and forgetting) on each subject. Table 3 compares the average incremental model performance over all subjects to examine the effectiveness of incremental strategies. Performance comparison between different IL strategies are discussed in following Section 5.2.1 and Section 5.2.2.

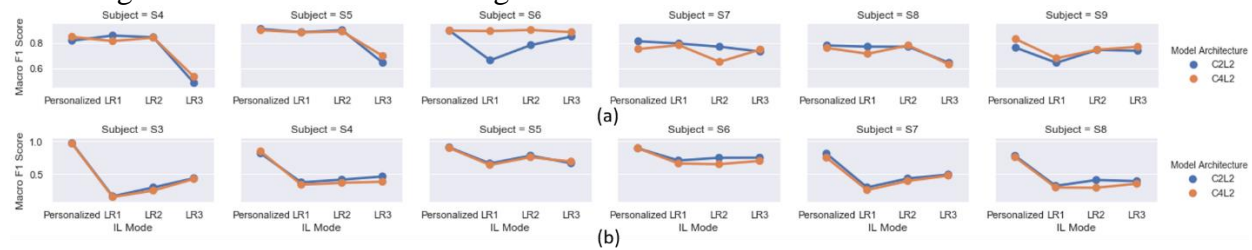

Figure 8. OtO IL Results. (a) Incremental learning performance on new subject; (b) Forgetting effect on the current subject after the model's adaptation to a new subject. S3 has no incremental performance due to the lack of the precedent; S9 has no forgetting performance as it is the last subject. The dots were the average Macro F1 Score over five-round SRS. The personalized models in (a) and (b) were trained from scratch for the current subject.

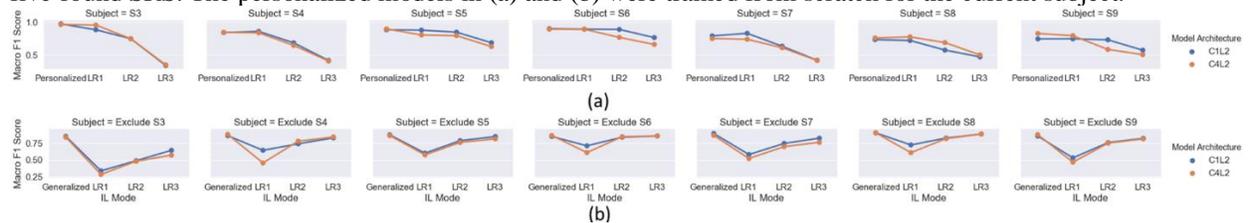

Figure 9. MtO IL Results. (a) Incremental learning performance on the new subject; (b) Forgetting effect on the rest subjects after the model's adaptation to a new subject. The dots were the average Macro F1 Score over five-round SRS. Personalized (a) and generalized (b) models were trained from scratch for a single subject or group of subjects.

Table 3. Evaluation of Incremental Learning. Average Macro F1 Score across subjects was used as metric.

| Incremental Training Strategies | | | Incremental Performance | | Forgetting Performance | | Baselines | |
|---|---|---|---|---|---|---|---|---|
| | | | Macro F1 | Change[*] | Macro F1 | Change[**] | Personalized on Target Subject | Generalized on Rest Subjects |
| OtO | C2L2 | LR1 | 0.808 | -2.4% | 0.422 | -49.0% | 0.828 | N/A |
| | | LR2 | 0.812 | -1.9% | 0.516 | -37.7% | | |
| | | LR3 | 0.682 | -17.6% | 0.535 | -35.4% | | |
| | C4L2 | LR1 | 0.793 | -4.3% | 0.393 | -52.6% | 0.829 | N/A |
| | | LR2 | 0.815 | -1.7% | 0.454 | -45.2% | | |
| | | LR3 | 0.710 | -14.4% | 0.507 | -38.8% | | |
| MtO | C1L2 | LR1 | 0.831 | -1.0% | 0.589 | -32.2% | 0.839 | 0.868 |
| | | LR2 | 0.730 | -13.0% | 0.739 | -14.8% | | |
| | | LR3 | 0.523 | -37.6% | 0.814 | -6.3% | | |
| | C4L2 | LR1 | 0.829 | -2.4% | 0.503 | -42.1% | 0.849 | 0.868 |
| | | LR2 | 0.691 | -18.5% | 0.732 | -15.7% | | |
| | | LR3 | 0.494 | -41.8% | 0.791 | -8.9% | | |

[*]The "Change" column represents the performance difference between the personalized model (trained from scratch) as baseline and the incremental CLN model for the targeted subject.

\*\*For OtO scheme, the "Change" column represents the performance difference between personalized model (trained from scratch) as baseline and incremental CLN model for the preceding subject. For MtO scheme, the column represents the performance difference between the generalized model (trained from scratch) as baseline and the incremental CLN model when being tested on the rest subjects.

### 5.2.1 Incremental Performance

In terms of model complexity, results in Figure 10-a show increasing the convolutional layer depth under OtO scheme did not consistently improve the incremental performance. The "shallow" CLN architecture tended to achieve a higher performance under MtO scheme. The observed superior incremental performance of the shallow CLN model can be explained by the nature of features learnt from shallower convolutional layers. The "coarse" features learnt from shallower layers has higher generality and transferability across subjects conducting similar tasks [19, 60], which allows the shallow CLN model adapts to target subject by effectively extracting generic features. Besides, the shallow architecture with reduced complexity can regularize the deep model, which in turn improves the model generality on *target*. It is also important to note the nature of data used for model training. The deeper model is suitable for processing large-scale image data [17]. E.g., even the tiny version ImageNet dataset contains 100,000 images across 200 classes. In this study, one subject's dataset contained around 1,000 "Motion Images" with less than six classes of postures to be recognized. The dataset's simplicity may eliminate the need for an overly complex model.

Figure 10-a,b shows decreasing the LR can impede the effective IL. The MtO scheme, in particular, required a larger LR (LR1) than the OtO to achieve optimal incremental performance (see Figure 10-b), regardless of model complexity. This can be explained by that the generalized CLN model used under MtO scheme required a larger extent of model weight updating when adapting to a new subject. The performance of optimal MtO incremental model (C1L2+LR1, Macro F1 Score-0.831) degraded by only an average of 1.0% from the baseline model on *target*. Such incremental performance outperformed that achieved under the optimal OtO incremental model (C4L2+LR2, Macro F1 Score-0.815).

The higher incremental performance from the MtO scheme can be attributed to the generalized CLN model used. Higher adaptability has been achieved when transferring from a domain comprising activities with rich variability to a domain with lower variability in the previous study [19], which suggests the convolutional features should be ideally trained on datasets with complex sets of activities. In this study, the generalized CLN model was trained from more subjects' postures with higher variability. The generalized dataset can reduce both bias and variance for developing a recognition model, which improves the model's adaptability to the *target*. These may also explain the observation that the MtO incremental model can even occasionally outperform personalized models (e.g., S4, S7, and S9 in Figure 11-a). The results suggest that the proposed CLN model under MtO scheme has the high potential for "adaptive personalization" [11].

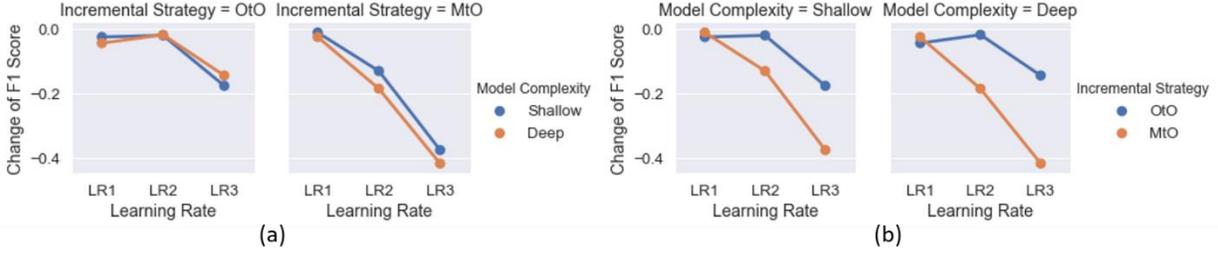

Figure 10. Incremental Performance Comparison. The "Change in F1 Score" is the based on the column "Change" under "Incremental Performance" in Table 3.

### 5.2.2 Forgetting Performance

Intuitively, the deep models should have a better control of forgetting effect, considering their higher learning capacity (for retaining the memory of learned information). However, results in Figure 11-a showed shallow CLN models tended to outperform the deep in controlling forgetting effect, regardless of IL strategies and LR. One explanation is that features learnt from shallower convolutional layers are transferable and can be generalized across subjects [19]. In this sense, the shallow CLN model can still capture generic features from *source* subjects after adapting to the *target*. It is also worth noting "fine" features learnt from deeper convolutional layers tend to be more personalized and subject-specific [19, 60]. A deeper architecture was prone to overfitting the *target* with small-size training data during adaptation in this study, which resulted in a high forgetting effect on learned subjects. Additionally, the shallow architecture, with fewer parameters, tends to regularize CLN models and control the forgetting effect.

The MtO scheme achieved higher performance in controlling forgetting effect than the OtO scheme regardless of LR and model complexity, as shown in Figure 11-b. Generic features learned from a generalized CLN model can contribute to controlling catastrophic forgetting. The forgetting can be attribute to the "Concept Drift" [61], where the posture distribution of a new subject is different from that of learned subjects in this study. The forgetting performance degraded significantly when the new subject had new posture classes (see forgetting effect on S3 in Figure 8-b). When the posture dataset is constructed from a larger group of subjects, it may reduce the extent of posture distribution difference.

Figure 11 shows the LR can effectively control the forgetting effect. The forgetting effect (denoted by Change of F1 Score from baseline) reduced consistently when decreasing LR (from LR1 to LR3) regardless of model complexity and learnings schemes. A lower LR restricted CLN models' adaptation to *target*, thus controlling the forgetting of learnt postures from *source*.

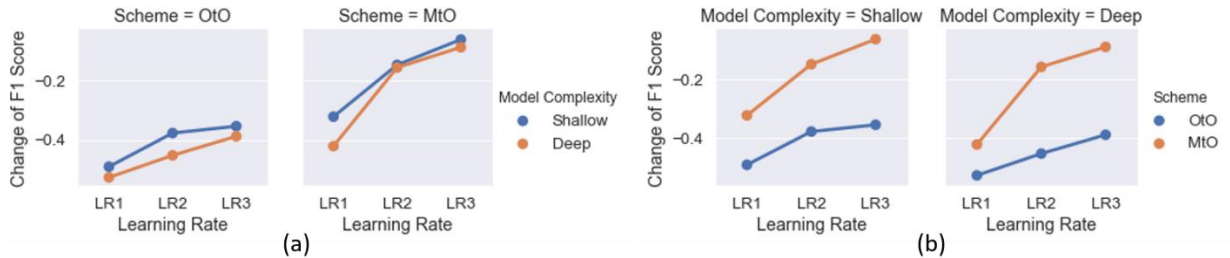

Figure 11. Forgetting Performance Comparison. The "Change in F1 Score" is the based on the column "Change" under "Forgetting Performance" in Table 3.

In summary, the proposed CLN model with shallow convolutional layers (C1L2) under MtO scheme emerged as an appropriate IL model, considering higher incremental and forgetting performances. It is worth noting that when the LR (i.e., model adaption speed) decreased, incremental and forgetting performances showed a contradiction for a given incremental CLN model. There is still a potential to balance the performances by tuning the LR. Particularly, the incremental CLN model under C1L2+MtO and LR2 balanced incremental and forgetting performances, with an average performance degradation of -13.0% and -14.8% from baselines, respectively. The model with the C1L2+MtO+LR2 incremental strategy was, therefore, applied as the incremental CLN model for further evaluation in subsequent sections.

### 5.3 Evaluation of Applying Incremental CLN Model for Posture Recognition

This section examined the recognition performance for each type of postures under identified incremental model (with the strategy of C1L2+MtO+LR2. Specifically, Section 5.3.1 investigates i) what kind of workers' postures can be incrementally learnt from new data; and ii) what kind of postures can be remembered after the model adapted to new postures. In addition, this study also identified the effective sensor placement and channels for recognizing workers' posture when using the incremental model in Section 5.3.2.

#### 5.3.1 Incremental Learning and Forgetting of Postures

We compared the CLN model's performance degradation from baseline models on each posture, where a smaller performance degradation (measured by "Change of F1 Score" in Table 4) denotes a higher incremental/forgetting performance for the given posture. Table 4 shows, when applying the incremental CLN model on target, the average performance degradation (across subjects S3-S9) for recognizing WO, WK, KN, KN, ST, SQ, and BT was within 5.8% from the baseline; while the incremental performance on MO was reduced by 34.8%. Table 4 also shows, when testing the incremental CLN model on source subjects, the average performance degradation for recognizing WO, WK, KN, KN, ST, SQ, and BT was within 12.5% from the baseline; whereas the forgetting performance on MO was reduced by 33.8%.

The results suggest most tested postures (WO, WK, KN, KN, ST, SQ, and BT) has the potential of being learnt incrementally and remembered by the incremental CLN model. This may indicate these postures in construction tasks have less inter-person variability among workers. Notably, the posture MO was "hard-to-learn" and "easy-to-forget" under IL. One possible explanation is that MO (postures related to climbing up/down) had higher variation and idiosyncrasy among workers, which lacks the adaptability across subjects. In addition, the posture MO showed only in two subjects (S5 and S8 in Table 2) and accounted for least proportion of postures from each subject. The lack of training data can also limit the recognition of MO under IL.

#### 5.3.2 Identifying Effective Sensor Placement and Channels

Figure 11 shows the influence of each sensor channel group on each posture with respect to incremental (a) and forgetting (b) performances, respectively. The performance degradation was measured by the change in F1 Score from baseline incremental CLN model after permuting each group of sensor channels.

Recognition performance degraded consistently regardless of which sensor channel group was permuted. Such results reaffirm the finding that a full-body sensor placement tends to achieve higher recognition performance in related study [24]. However, sensor channels of different nature impose varying influence on posture recognition under IL. For each posture, we ranked sensor channel groups by their change in F1 (from high to low) through averaging incremental and forgetting performances after permutation. The top 3 groups of "influential" channels are summarized in Table 3.

Table 4 shows most of the identified influential sensor channels for various postures were from accelerometer, except for the posture WK. Such results suggest that motion data from accelerometers contribute more for charactering workers' postures in daily tasks in this research context. Similarly, the effective sensor placement with high generality regarding each posture can also be identified. E.g., calf is influential in charactering lower-body awkward postures, like KN and SQ; while chest is appropriate for detecting upper-body postures, such as BT. Given the influential channels were identified considering both incremental and forgetting performance, their placement also indicates the appropriate body parts for discriminating works postures with similarity and charactering same postures across subjects under the naturalistic condition. Knowing such information can guide the effective sensor placement, which can benefit the use of DNN-based recognition models by further reducing the required number of sensors and computational load [62], without impairing the model performance.

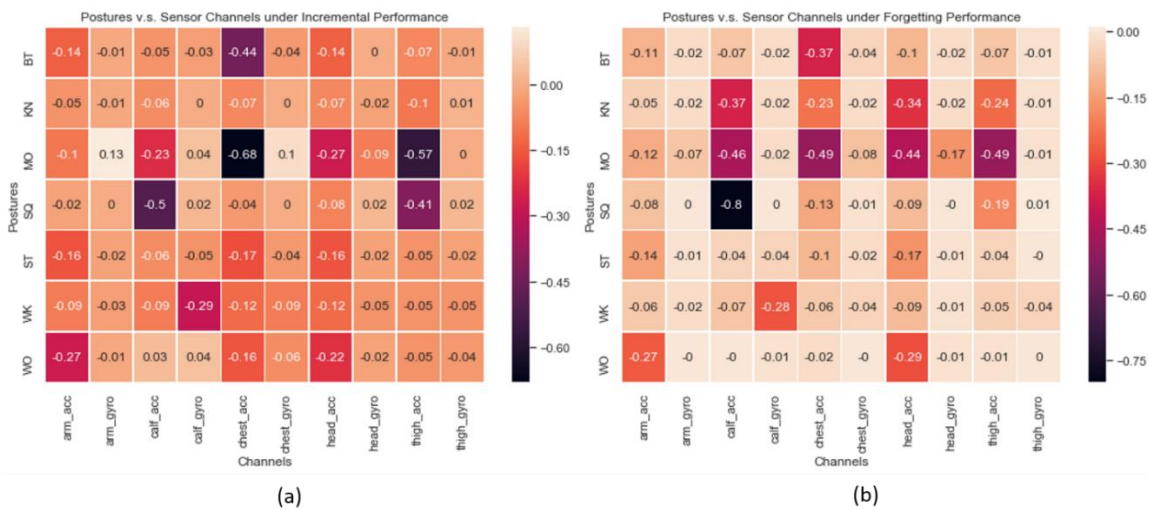

Figure 12. Influences of Sensor Placement and Channels on Posture Recognition under IL: (a) incremental performance and (b) forgetting performance. The value in each cell is the average across subjects (S3-S9 in the generalized dataset).

Table 4. Rank of Influential Sensor Channels

| Postures | Top-3 Influential Channel Groups (Change in F1 Score after permutation) | | |
|---|---|---|---|
| BT | chest_acc (-41%) | arm_acc (-13%) | head_acc (-12%) |
| KN | calf_acc (-21%) | head_acc (-21%) | thigh_acc (-17%) |
| MO | chest_acc (-58%) | thigh_acc (-53%) | head_acc (-35%) |
| SQ | calf_acc (-65%) | thigh_acc (-30%) | head_acc (-9%) |
| ST | head_acc (-17%) | arm_acc (-15%) | chest_acc (-13%) |
| WK | *calf_gyro* (-29%) | head_acc (-11%) | chest_acc (-9%) |
| WO | arm_acc (-27%) | head_acc (-25%) | chest_acc (-9%) |

## 5.4 Evaluation of Posture-based MSDs Risk Assessment

This section evaluates the ergonomics assessment based on postures recognized from the proposed incremental CLN model, with identified IL strategy "C1L2+LR2+MtO" in Section 5. Given the posture misclassification on *target* subject from the incremental model (see Figure 13), we examined the MSDs risk assessment result based on the use of postures from Ground Truth (G) and such Incremental (I) CLN model in Table 5.

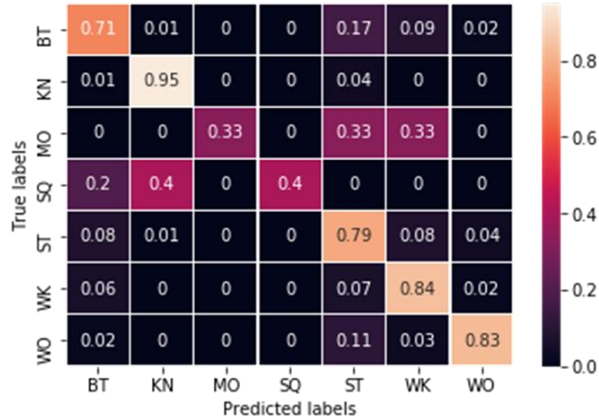

Figure 13. Confusion Matrix for the Incremental CLN Model in Quasi-Experiment. The confusion matrix was based on the combined posture recognition result from the test dataset of each subject in the generalized dataset (S3-S9). The values in the matrix were normalized for better visual interpretation. The model's incremental performance (on generalized dataset) was: Macro F1 Score-0.708, Accuracy-0.812.

Table 5. Comparison of Posture-based MSDs Risk Assessment.

|    | Count of MHT* Breach | | Total Duration of MHT Breach (s) | | Detected MHT Time (s) | | Frequency (Times/Min) | | Posture Proportion (Percentage) | |
|----|-----|-----|------|------|-----|-----|-----|-----|------|------|
|    | G** | I** | G    | I    | G   | I   | G   | I   | G    | I    |
| BT | 5   | 5   | 35.5 | 27.5 | 15  | 9   | 4.0 | 5.1 | 11.1 | 12.3 |
| KN | 6   | 5   | 50.5 | 48   | 13  | 15  | 0.6 | 0.7 | 8.8  | 8.9  |
| MO | 0   | 0   | 0    | 0    | 1.5 | 1   | 0.2 | 0.2 | 0.3  | 0.3  |
| SQ | 0   | 0   | 0    | 0    | 2   | 1.5 | 0.2 | 0.1 | 0.4  | 0.2  |
| ST | 0   | 1   | 0    | 30.5 | 17  | 31  | 5.6 | 8.7 | 31.9 | 32.1 |
| WK | 0   | 0   | 0    | 0    | 13  | 8   | 3.8 | 5.3 | 14.5 | 16.9 |
| WO | 18  | 15  | 168  | 124  | 30  | 21  | 3.6 | 4.8 | 33.0 | 29.4 |

*The MHT thresholds described in Section 4.5 were scaled down to 10% of original values, given the test dataset was the 10% subsample of one subject's motion dataset.
** G and I combined ground truth and recognized postures from the test dataset of each subject in the generalized dataset (S3-S9).

shows how the posture recognition errors influence the MSDs assessment results. The incremental model made misdetections for KN and WO breaching the MHT thresholds, despite the relatively high recognition accuracy of KN and WO (see Figure 13). The total duration of awkward postures (BT, KN, and WO) breaching MHT was also underestimated when using the incremental model. The underestimation was caused by misclassifying the awkward postures (BT, KN, and WO) as normal postures, particularly ST, as shown in Figure 13. The misdetection of MHT breach also result from the recognition errors that occurred when detecting continuous postures. E.g., Figure 14 shows misclassifications occurred around 78s and 86s when recognizing continuous KN from S3. Multiple misclassifications also occurred between 70s and 84s when detecting continuous WO from S7. The interruption led to the underestimated posture holding time and overestimated posture repetitiveness (see Frequency in ). Additionally, the incremental CLN models were prone to errors when detecting the beginning (24s in Figure 14-a) and ending (2s and 36s in Figure 14-a) of continuous postures. Misclassifications between KN and ST at posture transitions may be explained by the inter-class similarity between consecutive postures.

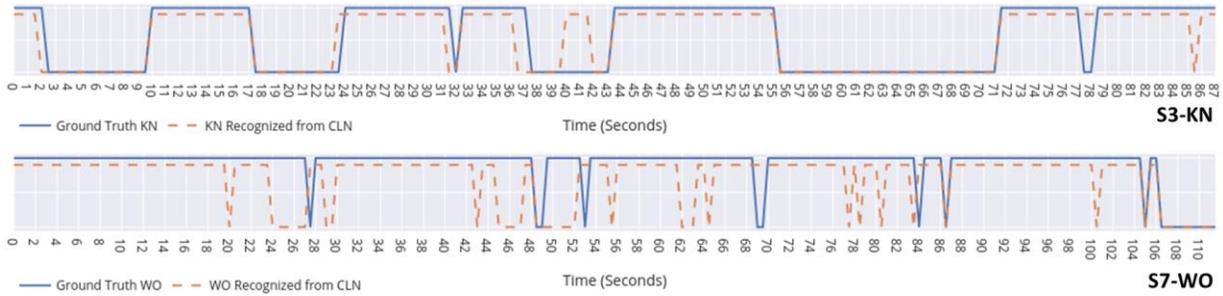

Figure 14. Recognition Errors on Test Datasets. Using KN from S3 and WO from S7 as an example.

The awkward posture proportions are input for OWAS rules when determining MSDs risk levels. shows the incremental models provided a close posture proportion estimation to the ground truth over all subjects. Table 6 presents the posture proportion for each subject with associated risk levels. The results show most of the risk levels were correctly identified based on the use of recognized postures with proportion-based thresholds. It is also worth noting misdetection of BT from S5 resulted in underestimated risk levels. False alarms also emerged for S8, where the risk level of WO was overestimated, which might be caused by misclassifying ST as WO.

Table 6. Ergonomics Risk Level (see Section 4.5.1 for detail) based on Recognized Awkward Postures.

|  |  | S3 | S4 | S5 | S6 | S7 | S8 | S9 |
|---|---|---|---|---|---|---|---|---|
| BT | G | 15.4%-I | 11.2%-I | 21.1%-II | 9.9%-I | 0.0%-I | 15.4%-I | 12.3%-I |
|  | I | 16.0%-I | 14.7%-I | 19.3%-I | 8.1%-I | 4.7%-I | 15.4%-I | 12.3%-I |
| KN | G | 57.1%-III |  |  |  |  |  |  |
|  | I | 58.3%-III |  |  |  |  |  |  |
| SQ | G | 2.9%-I |  |  |  |  |  |  |
|  | I | 1.1%-I |  |  |  |  |  |  |
| WO | G |  |  | 37.3%-II | 65.2%-III | 95.8%-III | 15.4%-I | 0.0%-I |
|  | I |  |  | 38.6%-II | 52.8%-III | 83.1%-III | 20.5%-II | 0.6%-I |

Results in this section suggest that using the recognized postures for proactive MSDs risk assessment shows promising results, despite the errors from proposed incremental CLN models. It is important to note the incremental CLN model tended to make errors when detecting continuous awkward postures and postures at transition. The misclassification affected the detection of postures breaching MHT thresholds. In addition, there was also a tendency of underestimating the exposure to awkward postures due to misclassifying awkward postures as normal ones. One penitential improvement for recognizing awkward postures is implementing a higher penalty for misclassifying awkward postures as natural postures (e.g., ST and WK) during model training. One can also shorten the recommended MHT thresholds as a compensation, considering the errors from recognition models.

# 6   Conclusions

This study aims to leverage the incremental DNN models for monitoring and assessing awkward postures among construction workers, which ultimately contributes to the proactive prevention of MSDs-related injuries. The work discussed in this paper investigated the feasibility and strategies for applying the proposed CLN model under Incremental Learning using nine workers' naturalistic posture during daily tasks. The validity of applying ergonomics rules with recognized postures from the developed incremental CLN model was evaluated in a quasi-experiment.

Findings in Section 5.1 further validate the authors' previous observations [4] that Convolutional LSTM architecture with shallow convolutional layers has a high potential of automated posture recognition from workers motion data captured by wearable IMUs, which saves the efforts of manual feature engineering required by conventional ML models.

The systematic evaluation of IL strategies in Section 5 shows the feasibility of implementing the proposed CLN model for IL by direct incremental training. The generalized "shallow" CLN model ($C(64) \times 1 - RL(128) \times 2 - Sm$) using LR1 ($-10^{-2}$ in Adam optimizer) under MtO learning scheme is a promising IL strategy when adapting to *target* subject, which achieved comparable recognition performance to the baseline personalized model on *target*. By tuning down the LR, the configured incremental CLN model can potentially balance the adaptation on postures of target subject and forgetting effect on learnt postures from source subjects. These findings suggest the proposed CLN-based recognition model has high potential of "adaptive personalization".

The detail examination of incremental model performance in Section 5.3.1 shows most of the tested postures (e.g., WO, WK, KN, KN, ST, SQ, and BT) had less inter-subject variability, which allows the incremental model to learn adaptatively and with controlled forgetting. It is also worth noting postures with high inter-subject variation and fewer data for training are hard to be adapted to and prone to be forgotten. Results in Section 5.3.2 suggest the effective sensor placements for charactering workers' postures in construction tasks. For example, chest and calf emerged as appropriate sensor placements for detecting upper-body and lower-body postures, respectively. A selective sensor placement can help reducing the computational complexity of DNN models while minimizing model performance degradation.

Applying the ergonomics rules (OWAS and MHT) on recognized and ground truth postures yields comparable injury risk assessment results. Findings in Section 5.4 indicate the proposed incremental CLN model can provide reliable results under the posture-based MSDs risk assessment. However, it is also important to note i) the awkward postures tended to be misclassified as normal postures and ii) continuous awkward postures were prone to be interrupted. Both recognition errors can lead to the potential underestimation of MSDs-related risks.

Applying the ubiquitous wearable sensing for automated jobsite safety monitoring is becoming an emerging trend in construction. The developed incremental DNN-based model in this study contributes to the workers' awkward posture detection using wearable IMUs sensors through: i) automated motion feature learning for high-performance recognition and ii) continuously learning new postures while retaining the memory for learned postures in the real-world scenarios. Beyond posture recognition, this study also shows the recognized postures can be linked to the ergonomics risks by using posture-based assessment rules. Timely MSDs risk detection can enhance workers' safety awareness, enable their self-correction, and prevent cumulative injuries in the long term.

## 7  Limitation and Further Works

The proposed incremental CLN model was devaluated with nine construction workers voluntarily joined this study. Current sample size is not enough to capturing varying commonly used awkward postures among workers' daily tasks. A well-designed sampling approach can be applied to collect more representative awkward postures from vulnerable worker groups, e.g., posture data related to BT among the masonry (considering the prevalent low-back pain of the masonry).

The learning capacity is limited when we used the DNN model under fixed architecture. Such limitation in our proposed model results in the challenge of achieving high adaptivity and controlling forgetting simultaneously. Further work can investigate an elastic model architecture to enlarge the model learning capacity.

Additionally, we gave equal importance for recognizing both awkward and normal postures. However, accurate detection of awkward postures should be the priority in this research context considering misdetections of awkward postures can result in underestimated ergonomics risks and escalated injury risk exposure. An attention mechanism can be adopted in further work, which allows the recognition model to focus more on targeted awkward postures during model training. These endeavours can further improve the performance of awkward posture detection and the validity of injury risk assessment in construction.

## Appendix

Table A. 1 Pseudo-Code for Awkward Posture Assessment

| MHT and Awkward Posture Proportion Assessment |
|---|
| postures ← list of postures (with timestamps) recognized |
| threshold ← pre-defined threshold for each targeted posture |
| result ← result of MSDs risk assessment in the unit time interval |
| pointer=0 ← pointer for counting consecutive postures |
| c ← number of posture classes recognized in the unit working time |
| n ← number of postures recognized in the unit working time |
| count ← count the continuous holding time of every posture |
| count[0,1]=1 ← initialize the first captured posture |
| sub_count ← buffer for saving holding time of specific posture |
| **for** i **in range**(n-1): |
|     **if** postures[i+1] equals postures[i]: |
|         count[pointer,1]+=1 |
|         count[pinter,0]=postures[i] |
|     **else**: |
|         pointer+=1 # the pointer moves to the next consecutive postures when a different posture comes |
|         count[poiner,0]=postures[i+1] |
|         count[pointer,1]=1 |
| count[:, column3]=(count[:, column2]+1)*0.5 # add a third column in count to record time of continuously held posture, the postures recognized under 50% overlap sliding window |
| **for** i **in range**(c): |
|     sub_count=count[column1 equals i and column2 greater than threshold[i]] # i breaching MHT threshold |
|     result[i, 1]= sub_count.length # total count of posture i breaching MHT |
|     result[i, 2]= sub_count.sum # total duration of posture i breaching MHT |
|     result[i,3]= count[column1 equals i].length/(count[column2].sum*60) # frequencies of posture i in 1 min |
|     result[i,4]= count[column1 equals i ].sum/count[column1].sum #proportion of posture i |
|     result[i,5]= count[column1 equals i].max # max holding time of posture i |
| **return** result |